\documentclass[conference]{IEEEtran}
\IEEEoverridecommandlockouts
\usepackage{cite}
\usepackage{amsmath,amssymb,amsfonts}
\usepackage{algorithmic}
\usepackage{algorithm}
\usepackage{graphicx}
\usepackage{textcomp}
\usepackage{xcolor}
\def\BibTeX{{\rm B\kern-.05em{\sc i\kern-.025em b}\kern-.08em
    T\kern-.1667em\lower.7ex\hbox{E}\kern-.125emX}}
\begin{document}

\title{OG: Equip vision occupancy with instance segmentation and visual grounding\\
}

\author{\IEEEauthorblockN{1\textsuperscript{st} Zichao Dong $\dagger$}
\IEEEauthorblockA{\textit{udeer.ai} \\
zichao@udeer.ai}
\and
\IEEEauthorblockN{1\textsuperscript{st} Hang Ji $\dagger$}
\IEEEauthorblockA{\textit{udeer.ai} \\
hang@udeer.ai}
\and
\IEEEauthorblockN{3\textsuperscript{rd} Weikun Zhang}
\IEEEauthorblockA{\textit{Zhejiang University} \\
zhangwk@zju.edu.cn
}
\and
\IEEEauthorblockN{4\textsuperscript{th} Xufeng Huang}
\IEEEauthorblockA{\textit{udeer.ai} \\
xufeng@udeer.ai}
\and
\IEEEauthorblockN{5\textsuperscript{th} Junbo Chen *}
\IEEEauthorblockA{\textit{udeer.ai} \\
junbo@udeer.ai}
\thanks{ 
$\dagger$ These authors contributed equally to this work.

Weikun Zhang was an intern at udeer.ai during the project.

* Corresponding author.

}
}

\maketitle

\begin{abstract}
Occupancy prediction tasks focus on the inference of both geometry and semantic labels for each voxel, which is an important perception mission. However, it is still a semantic segmentation task without distinguishing various instances. Further, although some existing works, such as Open-Vocabulary Occupancy (OVO), have already solved the problem of open vocabulary detection, visual grounding in occupancy has not been solved to the best of our knowledge. To tackle the above two limitations, this paper proposes Occupancy Grounding (OG), a novel method that equips vanilla occupancy instance segmentation ability and could operate visual grounding in a voxel manner with the help of grounded-SAM. Keys to our approach are (1) affinity field prediction for instance clustering and (2) association strategy for aligning 2D instance masks and 3D occupancy instances. Extensive experiments have been conducted whose visualization results and analysis are shown below. Our code will be publicly released soon. 
\end{abstract}

\begin{IEEEkeywords}
occupancy detection, semantic segmentation, instance segmentation, visual grounding
\end{IEEEkeywords}

\section{Introduction}
In recent years, perception for robotics has attracted significant attention since it is crucial for safety. Apart from that, human-robot interaction is also a promising area since it could simplify the usage of robots. As for perception tasks, occupancy is a vital modality that could make dense predictions for the entire physical scene. Various methods have been proposed to solve it. However, to the best of our knowledge, few attempts have been done for adding instance information in occupancy prediction. Similar to 2D semantic segmentation, instance tag is of vital importance for downstream perception which could split world basic elements (pixels for 2D) as separate entities. As a result, it is natural to inhabitant this property for occupancy detection. Speaking of human-robot interaction, various prompt methods including human clicking and text prompts are common interaction paths for humans to interact with the robot in perception tasks. Thus, a simple pipeline for human prompts as input and corresponding instance occupancy result is hungry for in the real world. It is worth noticing that instance information for diverse voxels is an essential part of achieving that target since occlusion is the devil in monocular occupancy detection. 

Firstly, in order to equip occupancy with instance label, additional clues should also be predicted. Though the two-stage from bounding box to instance mask pipeline is common in 2D instance segmentation, it is too heavy to fuse it in the current occupancy pipeline since there is no detection step for now. Instead, we predict an affinity field as a bridge to convert discrete voxels to instances. This method is implemented as a plug-and-play module which could be injected in any occupancy framework with trivial modification. 

Secondly, cooperation with Grounded-SAM is also challenging. Though SAM framework does well in responding human prompts, the result of SAM is in 2D modality. Considering each pixel could be projected as a scan line in physical world, there may be multiple 3D occupancy instances matching the 2D instance mask while satisfying the camera projection relationship. In order to tackle it, a clustering-based association matrix is designed. This method would take original occupancy geometry and semantic label, affinity field prediction result, and prior projection relationship knowledge into account. 

To the best of our knowledge, our proposed method is the first attempt to add instance segmentation task in occupancy prediction. Further, we are the first to accomplish occupancy grounding with satisfying result. We believe that our proposed method has the potential to contribute significantly to
the development of more robust and flexible 3D scene understanding methods, which can further benefit human-robot interaction.

\section{Related work}
\subsection{segmentation task}

Segmentation task contains semantic segmentation,instance segmentation and panoptic segmentation.
Semantic segmentation is a task which partition a image into regions with different semantic categories.
From FCNS\cite{long2015fully}, most deep learning-based methods formulate Semantic segmentation as per-pixel classification,applying a classification loss to each output pixels. Modern methods like DANet\cite{fu2019dual}, OCNet\cite{yuan2021ocnet} and SETR\cite{zheng2021rethinking} focus more on how the network extracts better depth features. Different from Semantic segmentation,Instance segmentation
involves identifying and segmenting individual objects within an image. This means that multiple instances of the same object class can be different from each other based on their unique labels or masks. Mask classification is commonly used for instance segmentation tasks. 
Panoptic segmentation unifies semantic segmentation and instance segmentation, predicting a mask for each object instance or background segment.
MaskFormer\cite{cheng2021per} proposes mask classification is sufficiently general to solve both semantic and instance segmentation tasks in a unified manner using the same model, loss, and training procedure. 
Although segmentation tasks are widely applied, they still have limitations in certain fields, particularly in the domain of autonomous driving, where they fail to capture the geometric state of objects in 3D scene.

\subsection{Occupancy perception}

In recent years, Occupancy perception is essential in the field of computer vision, especially applied in the autonomous driving domain.
The task of occupancy prediction involves predicting the occupancy status and corresponding semantic labels for each voxel in a given 3D scene.
This task involves two types of inputs: 2D input and 3D input. The input of 3D contains the input of lidar and multiple cameras, containing a wealth of spatial information, but equipment is expensive. The input to 2D is a monocular camera,lack of depth information,but equipment is cheap. MonoScene\cite{cao2022monoscene} is a occupancy perception framework which
outputs 3D occupancy perception when given a monocular camera input. The framework incorporates a module for 2D feature line of sight projection (FLoSP) to link the 2D and 3D U-Net and a 3D context relation prior (CRP) layer to enhance contextual learning. OccDepth exploits implicit depth information from stereo images (or RGBD images) to help the of occupancy perception\cite{miao2023occdepth}. The framework uses Stereo-SFA module to better fuse 3D depth-aware features, and OAD module to produce depth-aware 3D features.
VoxFormer\cite{zhang2023occformer} is a transformer-based framework that operates in two stages. It starts by using sparse visible and occupied queries from a depth map and then propagates them to dense voxels using self-attention mechanisms. SurroundOcc\cite{wei2023surroundocc} fisrt proposes extract multi-scale features for each image and adopt spatital 2D-3D attention to lift them to the 3D volume space, then uses 3D convolutions to generate features on multiple levels. OpenOccupancy\cite{wang2023openoccupancy} propose the Cascade Occupancy Network (CONet) to refine the coarse prediction, which can enhances the occupancy perception performance.

\subsection{Visual grounding}

Visual grounding is a task aimed at accurately identifying the correspondence between prompts and corresponding objects or regions in an image. These prompts can be in the form of text, bounding boxes, or even manually labeled points on the image. The focus of this task is to establish precise connections between prompts and visual elements.
Grounded Language-Image Pre-training (GLIP) goes beyond the limitations of models constrained to a predefined set of fixed categories by utilizing textual information to enhance the effectiveness of detection models.
It unifies the visual grounding and detection tasks.
GroundDino\cite{liu2023grounding} proposes open-set object detector which marries Transfomer-based detector DINO with grounded pre-training,can detect arbitrary objects with human inputs such as category names or referring expression\cite{li2022grounded}.
Segment Anything Model\cite{kirillov2023segment} build a foundation model for segmentation which can use a rough box or mask,free-form text, or, in general, any information indication what to segment a image, unifies visual grounding and 2D segment tasks.
Grounded Segment Anything which is a interesting demo combines Grounding DINO and SAM to detect and segment anything with text,unifies visual grounding and 2D vision tasks\cite{li2022grounded,kirillov2023segment}.
Mask DINO\cite{li2023mask} is a unified transformer-based network structure that extends DINO by incorporating a segmentation component. This integration allows for the mutual enhancement of the detection and segmentation tasks.

\section{Proposed Method: OG}
\subsection{Background}\label{AA}

\textbf{Instance segmentation in occupancy} has great significance for the field of computer vision. The focus and challenges of instance segmentation compared to semantic segmentation are as follows: 
\begin{enumerate}
\item[\textbullet] Voxel-level precise segmentation: Instance segmentation needs accurately assign each voxel to its corresponding instance. This demand segmentation algorithms with advanced perception capabilities to segment instances of varying shapes, sizes, and orientations.
\item[\textbullet] Instance boundaries: Instance segmentation needs to overcome the hardship of boundaries between instances. Instances may exhibit overlaps, contacts, or occlusions, demanding accurate detection of instance boundaries and explicit segmentation.
\item[\textbullet] Instance association and contextual information: in instance segmentation we need to consider the association and contextual information between instances. Instances may exhibit semantic relationships and understanding such relationships can assist in accurate instance segmentation. Leveraging both local and global contextual cues is crucial for achieving high-quality results.
\end{enumerate}
Despite the achievements of prior methods, our work builds upon a combination of outstanding models to overcome the difficulties in instance segmentation. Next, we will review several important models employed in our research.

\textbf{MonoScene}\cite{cao2022monoscene} is a pioneering framework proposed for 3D Semantic Scene Completion (SSC). By using a single RGB image as input, MonoScene can generate a 3D voxel grid representation of the scene's geometry and semantics. This innovative framework surpasses prior approaches, achieving superior accuracy and comprehensive scene understanding.

MonoScene has three key stages. Firstly, a 2D convolutional neural network (CNN) extracts features from the RGB image, capturing crucial visual information for upcoming processing. The Second stage introduces the Features Line of Sight Projection (FLoSP), a 2D-to-3D feature projection module. FLoSP preserves spatial relationships during projection, enabling the subsequent 3D CNN to effectively leverage the projected information and comprehensively understand the scene's geometry and semantics. FLoSP plays a vital role in MonoScene's success. Furthermore, in the final stage, a 3D CNN refines the projected features and predicts the scene's geometry and semantics. 

\textbf{Grounded-SAM}, developed by IDEA-Research is a novel image tagging model that combines two existing models: Segment Anything Model (SAM)\cite{kirillov2023segment} and Grounding DINO\cite{liu2023grounding}.

\textbf{Segment Anything Model(SAM)}, proposed by the Facebook AI Research (FAIR) team, is currently the most influential model in the field of prompt-based image segmentation. This model is based on a transformer vision model, and it has been pre-trained with a masked autoencoder (MAE)\cite{he2022masked}.

SAM has three primary components: an image encoder, a prompt encoder and a mask decoder. The image encoder is based on a Vision Transformer (ViT)\cite{dosovitskiy2020image} and pre-trained with a masked autoencoder, which has been minimally adapted to process high-resolution inputs. This encoder operates once per image` before prompting the model. The prompt encoder handles two types of prompts: sparse (points, boxes, text) and dense (masks). Positional encodings\cite{tancik2020fourier} are used to represent points and boxes, while the text is embedded using an off-the-shelf text encoder from CLIP\cite{radford2021learning}. Dense prompts, i.e., masks, are embedded using convolutions and summed element-wise with the image embedding. The mask decoder efficiently maps the image embedding, prompt embeddings, and an output token to a mask. This design employs a modification of a Transformer decoder block\cite{vaswani2017attention} followed by a dynamic mask prediction head. The decoder block uses prompt self-attention and cross-attention in two directions (prompt-to-image embedding and vice-versa) to update all embeddings. After running two blocks, the image embedding is upsampled, and a multi-layer perceptron(MLP) maps the output token to a dynamic linear classifier, which then computes the mask foreground probability at each image location. SAM predicts multiple output masks and corresponding confidence scores for a single prompt to resolve ambiguity in prompts. 

\textbf{Grounding DINO} is an open-set object detector by marrying Transformer-based detector DINO\cite{zhang2022Dino} with grounded pre-training. Grounding DINO can detect arbitrary objects with interactive human inputs, such as category names or referring expressions. Grounding DINO utilizes a dual-encoder-single-decoder architecture, which includes an image backbone for image feature extraction, a text backbone for text feature extraction, a feature enhancer for image and text feature fusion, a language-guided query selection module for query initialization, and a cross-modality decoder for box refinement. 
The feature extraction part of Grounding DINO is an image backbone like Swin Transformer\cite{liu2021swin} and a text backbone like BERT\cite{devlin2018bert} to extract multi-scale image and text features. Then we feed these image and text features into the feature enhancer. The feature enhancer uses deformable self-attention for image features and vanilla self-attention for text features. It also has image-to-text and text-to-image cross-attention modules for feature fusion. Aiming to leverage the input to guide the object detection process, Grounding DINO's language-guided query selection module selects features more relevant to the input text as decoder queries. Each decoder query contains a content part that is learnable while training and a positional part formulated as dynamic anchor boxes.
The cross-modality decoder combines image and text features with an extra text cross-attention layer for better modality alignment. Grounding DINO also introduces a sub-sentence level text feature to avoid unwanted word interactions, enabling encoding of multiple category names while avoiding unnecessary dependencies among different categories.

\subsection{Instance segmentation via affinity field}

\begin{figure*}[htbp]
    \centering
    \includegraphics[width=16cm]{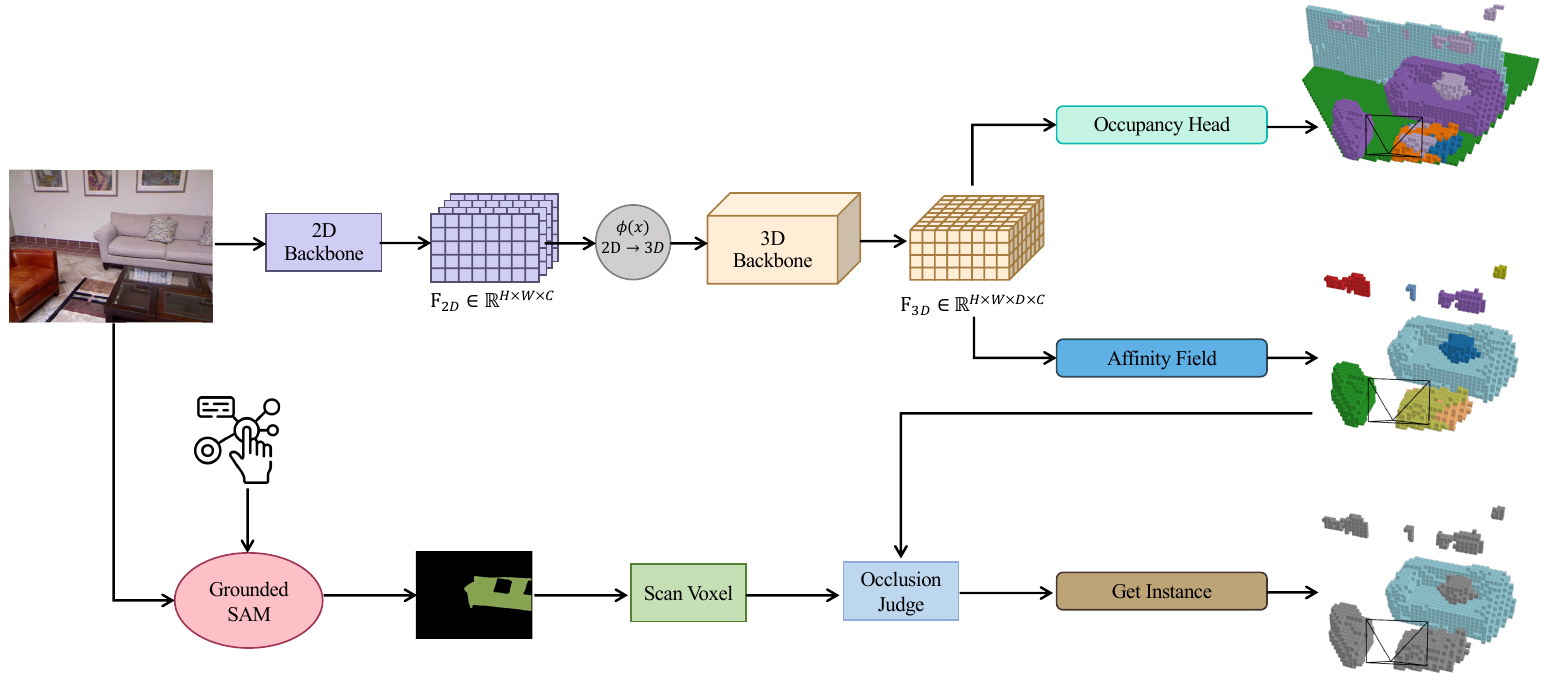}
    \caption{\small \textbf{Overall pipeline of OG}. Our framework not only enables the generation of occupancy but also provides instance segmentation through affinity fields. Additionally, we enhance the model with visual grounding ability by integrating Grounded-SAM. The first row displays the original occupancy result, where the color of voxels corresponds to the OVO~\cite{tan2023ovo}. The second row shows the instance segmentation result, where each instance is assigned a color that doesn't indicate semantics. The third row showcases the grounded results. We utilize the mask of "sofa" to obtain the occupancy of the sofa depicted in color, while the other instances are represented in gray.}
    \label{fig:model}
\end{figure*}

Fig.~\ref{fig:model} illustrates our overall framework. In order to acquire instance tag for each foreground voxel, we need extra clue to group voxels belonging to same instance in same cluster. Inspired by widely used affinity field in pose estimation, we formulate this voxel clustering problem as a affinity field regression problem. Firstly, we segment occupancy ground truth to separate instances by simple connected map segmentation algorithms. Given $n$ voxels belonging to a certain instance, we calculate the geometry center $(\bar{x}, \bar{y}, \bar{z})$ by getting the mean position of all voxels as  

\begin{equation}
\label{eq:center}
    (\bar{x}, \bar{y}, \bar{z}) = \left(\frac{1}{n}\sum_{i=1}^{n}x_i, \frac{1}{n}\sum_{i=1}^{n}y_i, \frac{1}{n}\sum_{i=1}^{n}z_i\right).
\end{equation}

The ground truth affinity value of each voxel $\mathcal{A}$ is defined as its original position minus the center of this instance. Note that for empty voxels, the ground truth value of corresponding affinity value is set as zero. We would use $\mathcal{M}$ to mask affinity loss from empty voxels during training. As for model architecture of affinity prediction head $\Psi$, it has the same structure as the segmentation head in monoscene with only change the final output number of channel to 3, matching delta of $(x, y, z)$. Mean square error is used for affinity field regression task, which is then added with original occuapncy losses.

\begin{equation}
    \mathcal{L}_\text{aff} = MSE \big(\Psi(F_{3D})\cdot \mathcal{M}, \mathcal{A}\big)
\end{equation}

\begin{equation}
    \mathcal{L}_\text{total} = \mathcal{L}_\text{ori} + \lambda \mathcal{L}_\text{aff}
\end{equation}

\subsection{Adapter between 23D}

The underlying principle of computer vision is that each 2D pixel in an image projection corresponds to a 3D ray originating from the camera's optical center and passing through that pixel. With knowledge of this 3D ray, we can locate the 3D point in space that projects to that pixel.
We can split the process of transforming 2D pixels into 3D rays into the following steps:
\begin{enumerate}
\item[\textbullet] Camera Model: The initial step involves defining the camera model, which establishes the relationship between the pixels in the image and the corresponding rays in 3D space. A camera model should have Extrinsic parameters and intrinsic parameters. Extrinsic parameters define the location and the orientation of the camera with respect to the world frame, and intrinsic parameters allow a mapping between camera coordinates and pixel coordinates in the image frame. The pinhole camera model, owing to its simplicity and effectiveness, is commonly employed for this purpose.
\item[\textbullet] Ray Direction: Following the camera model definition, the direction of the ray corresponding to each pixel is computed. This is achieved by utilizing the pixel coordinates in conjunction with the defined camera model.
\item[\textbullet] Ray Origin: The next step is determining the origin of the 3D ray, and it is usually the location of the camera in the 3D space. This origin is the starting point of the ray.
\end{enumerate}
These theories lead to an effective method for transforming 2D pixels into 3D rays, helping the process of 3D alignment. We propose the Pixel to Voxel Transformation and Clustering algorithm as a synthesis of the above-mentioned theories for our task.

\begin{algorithm}
\caption{Pixel to Voxels Transformation and Clustering}
\begin{algorithmic}[1]
\renewcommand{\algorithmicrequire}{\textbf{Input:}}  
\renewcommand{\algorithmicensure}{\textbf{Output:}} 
\REQUIRE 
Grounded SAM segmentation result;
MonoScene semantic result;
Affinity result;
Background list.
\ENSURE
Voxels with clustering information.
\STATE Initialize empty list for candidate voxels.
\FOR{each pixel in Grounded-SAM result}
    \STATE Transform pixel to scan-line voxels and append to candidate voxels
\ENDFOR
\STATE Initialize empty list for foreground voxels.
\FOR{each voxel in candidate voxels}
    \STATE If voxel's label is not in background list, calculate affinity and append voxel information to foreground voxels
\ENDFOR
\STATE Use a clustering algorithm (e.g., DBSCAN) to find clusters in foreground voxels with the help of affinity field information.
\STATE If multiple clusters are found, return the nearest one considering occlusion.
\end{algorithmic}
\end{algorithm}

This algorithm begins by taking the 2D segmentation result from the Grounded-SAM model as input. We traverse all pixels in the 2D segmentation result and transform each of them into a scan-line voxel. These scan-line voxels are saved in a candidate voxel list.
Afterthat, the algorithm receives the 3D semantic result and the affinity result from the MonoScene model. The semantic result is used to generate labels for each voxel in the candidate list, while the affinity result is used to calculate the center of each cluster.
The algorithm then iterates through each voxel in the candidate list. If the label of a voxel is not in the predefined background list, the algorithm calculates the affinity of the voxel and appends the voxel's information, including its coordinates, center coordinates, and label, to a foreground voxel list.
Finally, we apply a clustering algorithm such as DBSCAN to the foreground voxel list to get clusters. And if multiple clusters are identified, the algorithm will return the nearest cluster. This approach ensures the accurate and efficient transformation of 2D segmentation results into 3D voxel clusters.

\section{Experiments}
\subsection{Setup}
\textbf{Dataset}. We conduct experiments on \textbf{NYUv2}\cite{silberman2012indoor}, one of the most widely-used benchmarks for semantic occupancy prediction. The dataset consists of  1449 indoor scenes. Following the split setup in MonoScene, we utilize 795 scenes for training and reserve 654 scenes for inference at a scale of 1:4. The NYUv2 dataset consists of 12 classes.

\textbf{Implementation Details}. We kept the backbone parameters consistent with MonoScene except for adding the affinity prediction head. We train OG on the NYUv2 dataset for 50 epochs using 8 V100 GPUs.

\subsection{Main Results}
\begin{figure*}[htbp]
\centering     

\begin{minipage}{0.04\linewidth}
    \vspace{3pt}
    \rotatebox{90}{RGB Input}
\end{minipage}
\begin{minipage}{0.23\linewidth}
    \vspace{3pt}
    \centerline{\includegraphics[width=\textwidth]{figure/0462_1_color.jpg}}
\end{minipage}
\begin{minipage}{0.23\linewidth}
    \vspace{3pt}
    \centerline{\includegraphics[width=\textwidth]{figure/0271_1_color.jpg}}
\end{minipage}
\begin{minipage}{0.23\linewidth}
    \vspace{3pt}
    \centerline{\includegraphics[width=\textwidth]{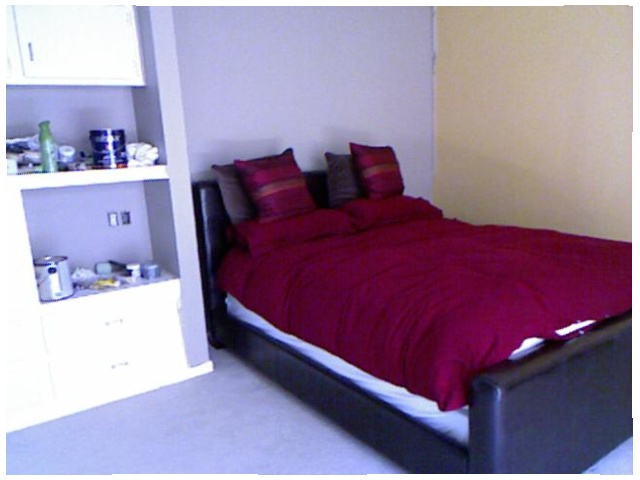}}
\end{minipage}
\begin{minipage}{0.23\linewidth}
    \vspace{3pt}
    \centerline{\includegraphics[width=\textwidth]{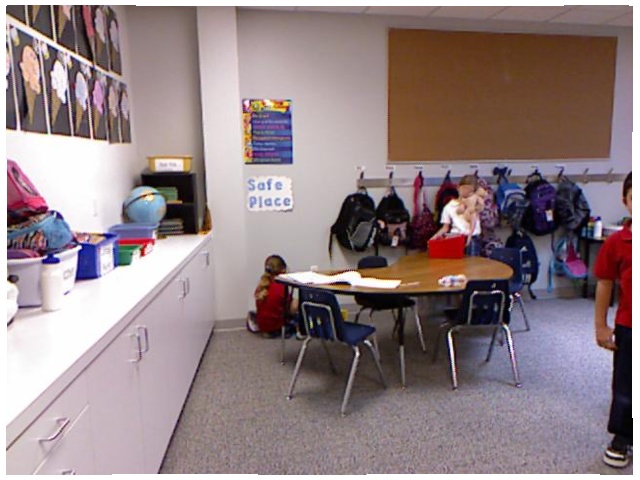}}
\end{minipage}

\begin{minipage}{0.04\linewidth}
    \vspace{3pt}
    \rotatebox{90}{Instance Seg.}
\end{minipage}
\begin{minipage}{0.23\linewidth}
    \vspace{3pt}
    \centerline{\includegraphics[width=\textwidth, viewport=1100 300 2300 1500,clip]{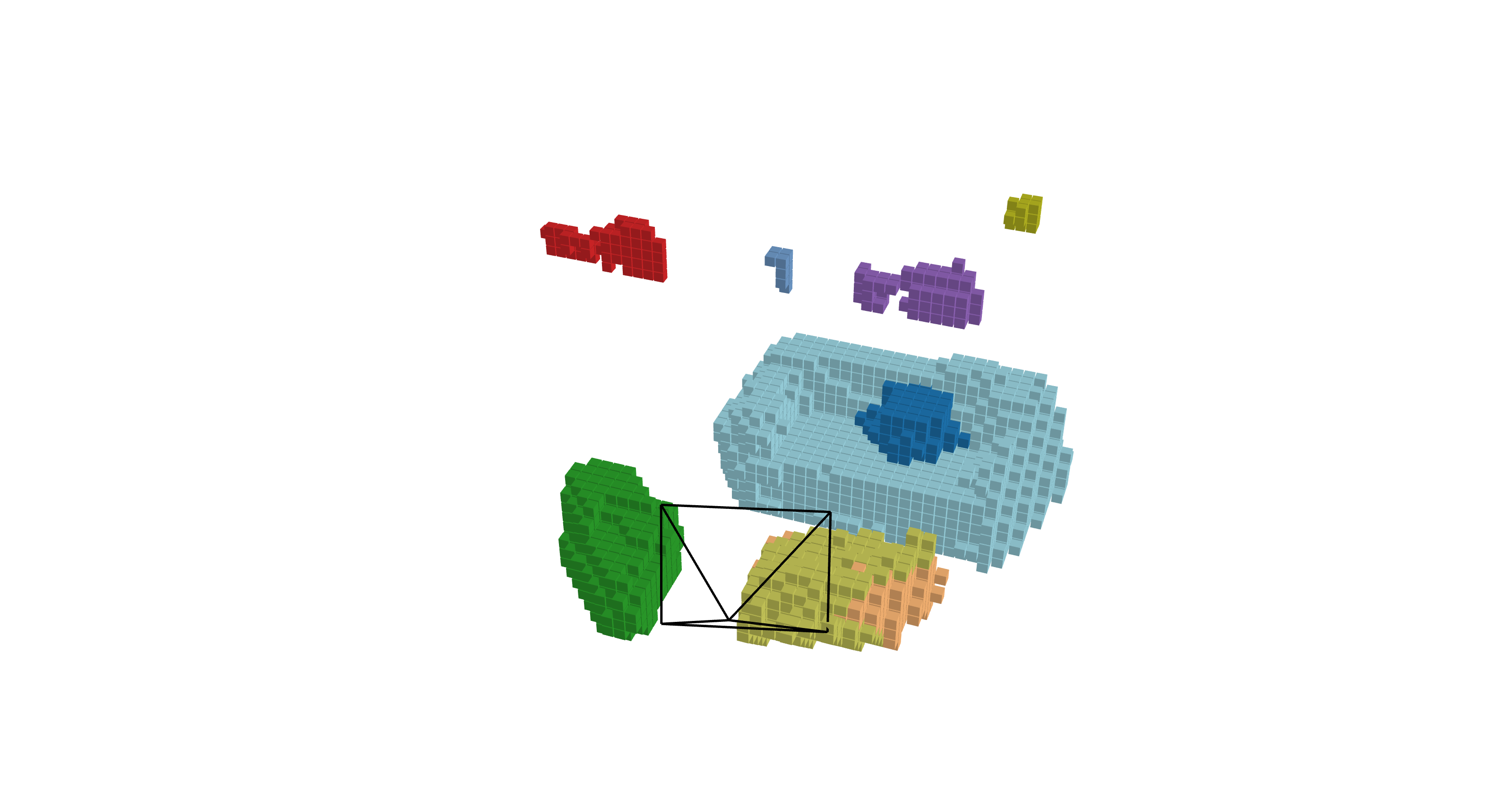}}
\end{minipage}
\begin{minipage}{0.23\linewidth}
    \vspace{3pt}
    \centerline{\includegraphics[width=\textwidth, viewport=500 350 1700 1400,clip]{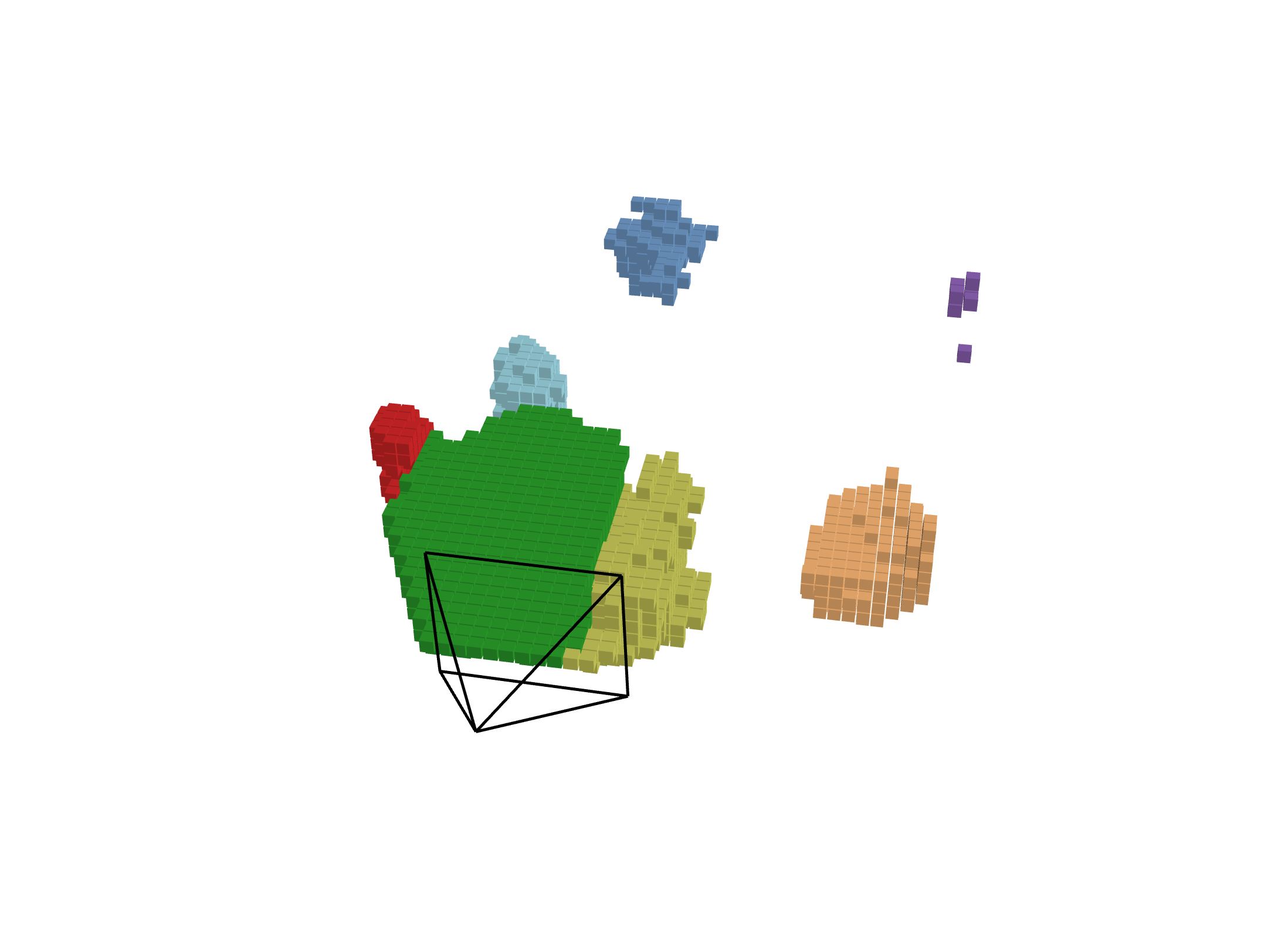}}
\end{minipage}
\begin{minipage}{0.23\linewidth}
    \vspace{3pt}
    \centerline{\includegraphics[width=\textwidth, viewport=600 150 2200 1600,clip]{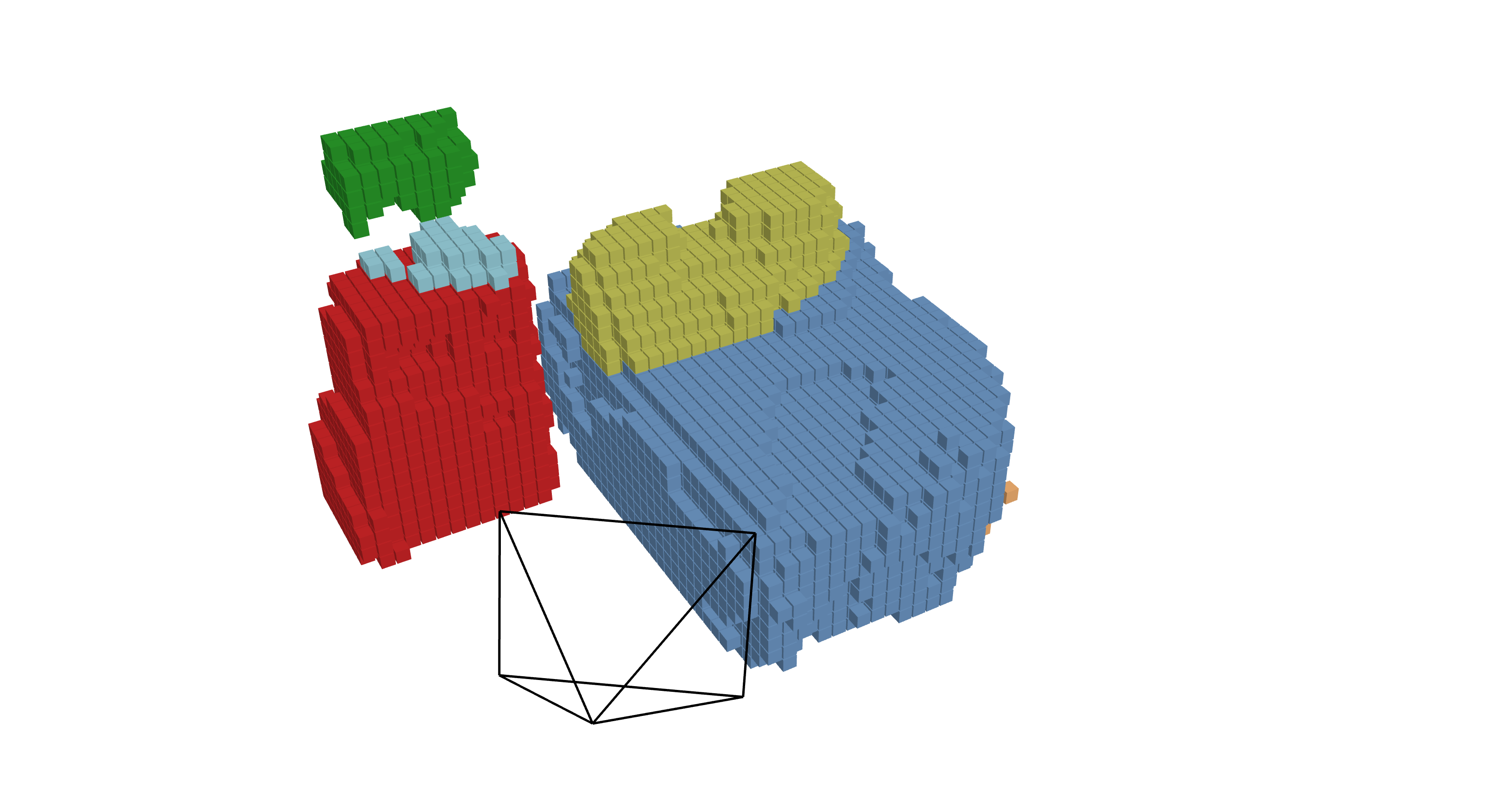}}
\end{minipage}
\begin{minipage}{0.23\linewidth}
    \vspace{3pt}
    \centerline{\includegraphics[width=\textwidth, viewport=850 300 2500 1600,clip]{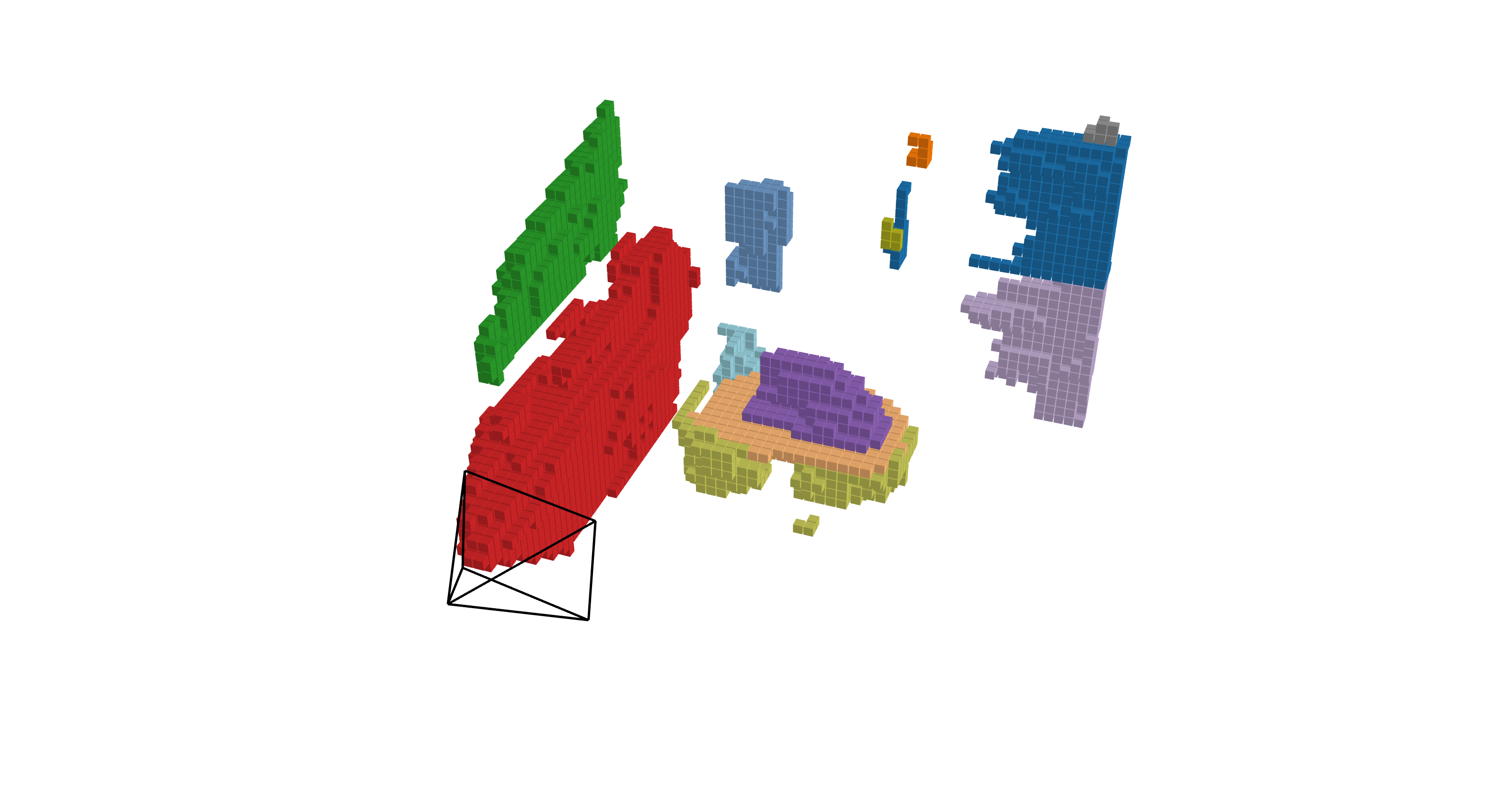}}
\end{minipage}

\begin{minipage}{0.04\linewidth}
    \vspace{3pt}
    \rotatebox{90}{GT}
\end{minipage}
\begin{minipage}{0.23\linewidth}
    \vspace{3pt}
    \centerline{\includegraphics[width=\textwidth, viewport=1100 300 2500 1500,clip]{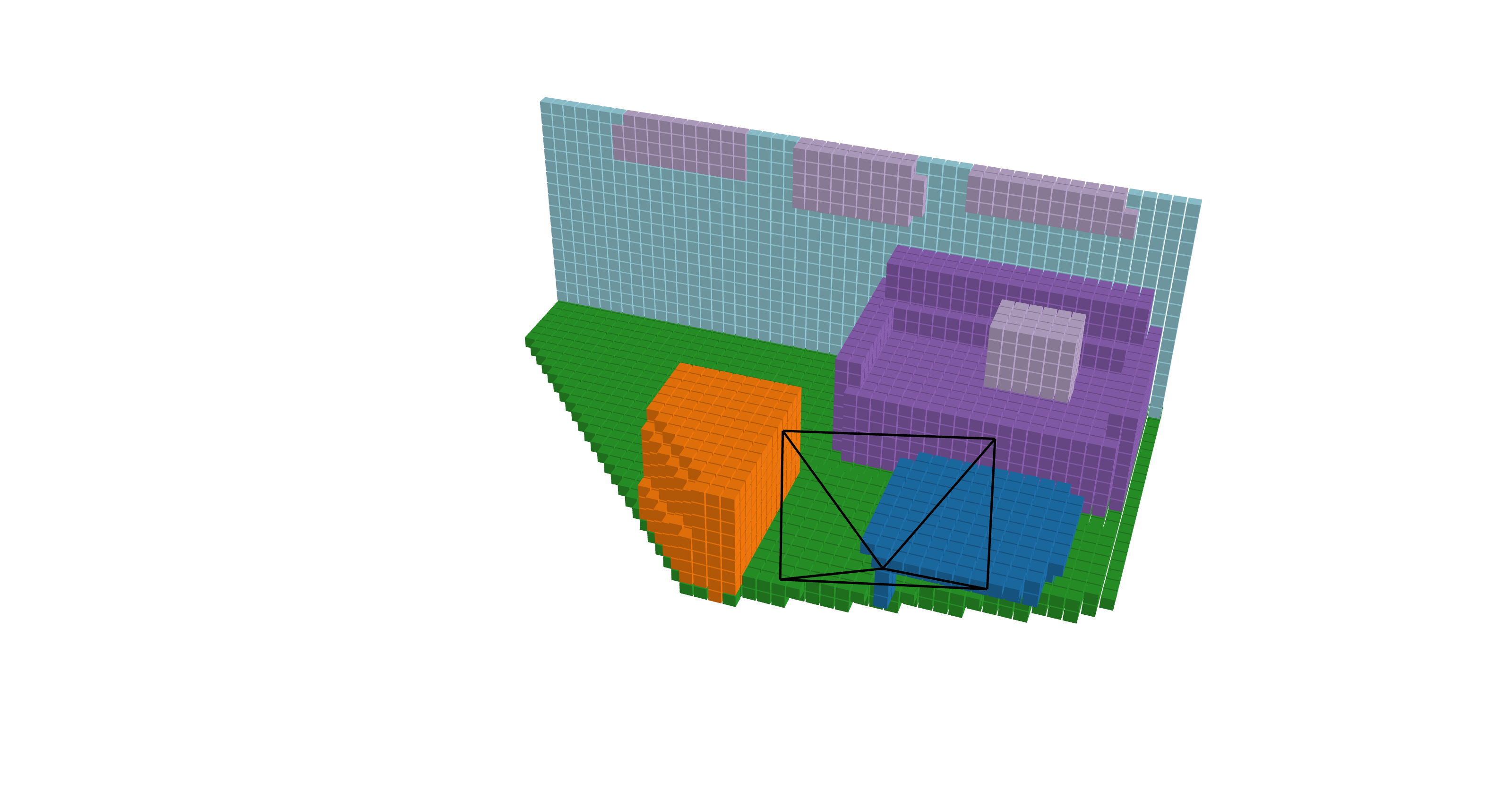}}
\end{minipage}
\begin{minipage}{0.23\linewidth}
    \vspace{3pt}
    \centerline{\includegraphics[width=\textwidth, viewport=500 250 1900 1500,clip]{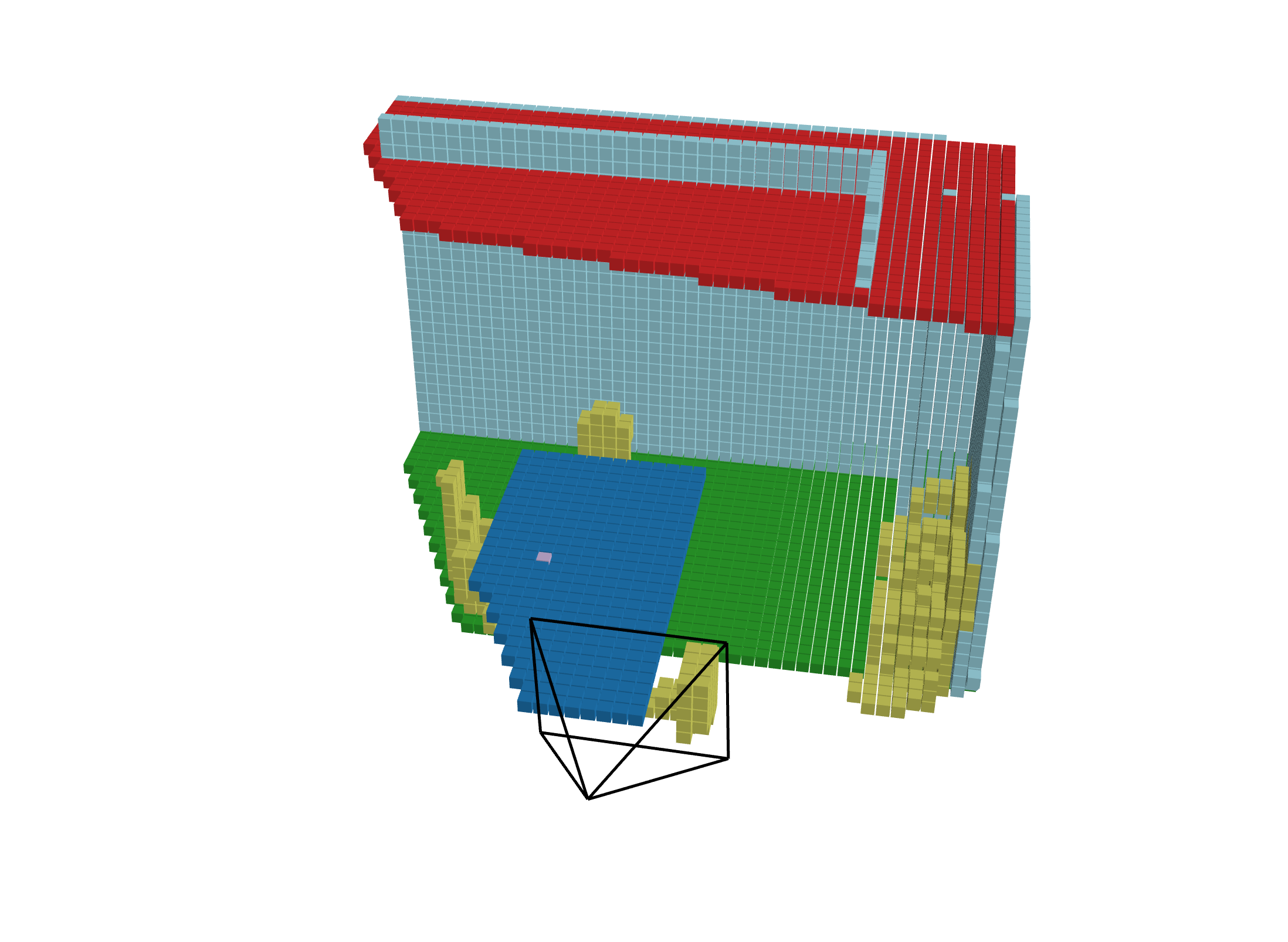}}
\end{minipage}
\begin{minipage}{0.23\linewidth}
    \vspace{3pt}
    \centerline{\includegraphics[width=\textwidth, viewport=600 0 2300 1700,clip]{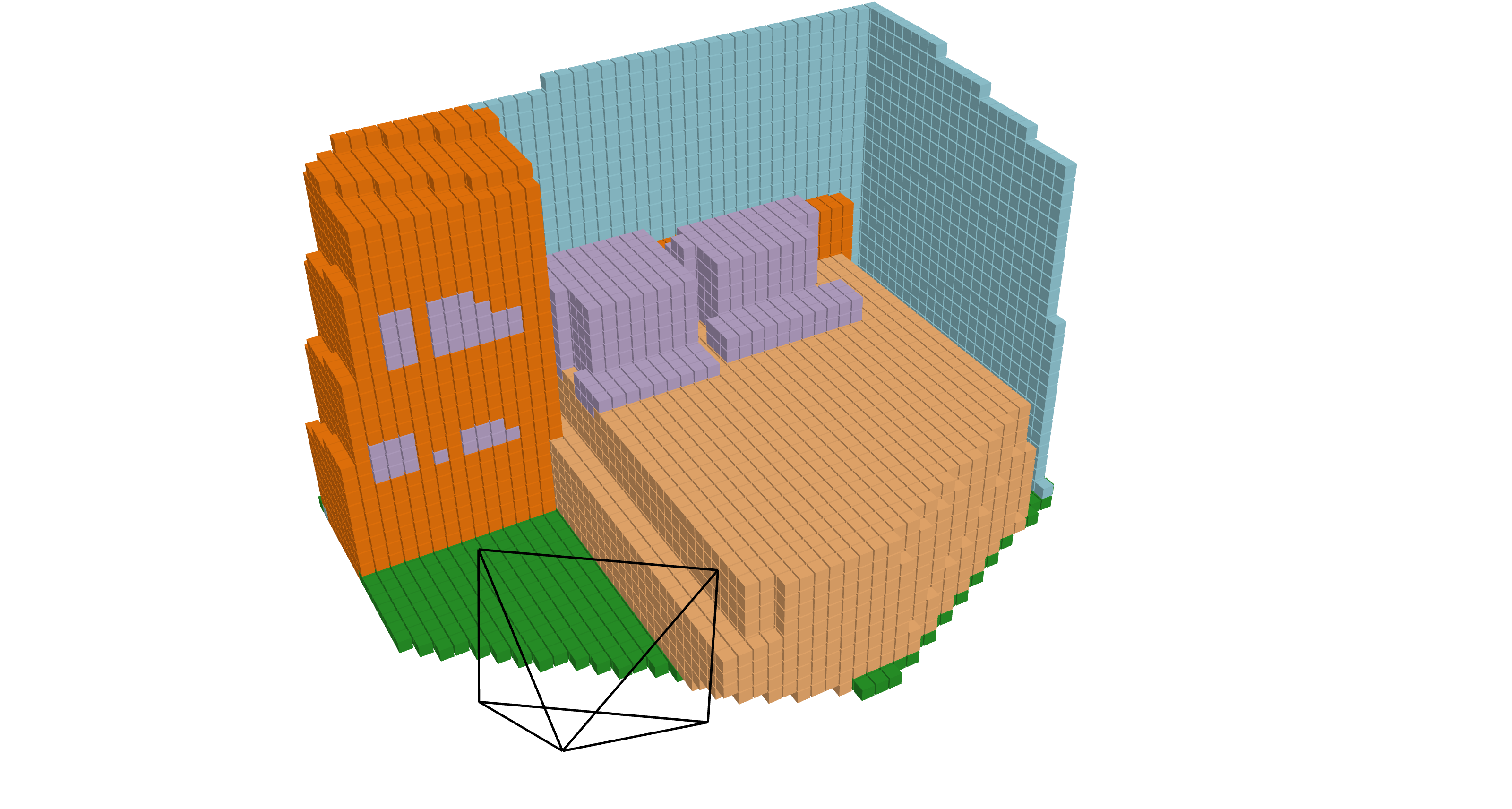}}
\end{minipage}
\begin{minipage}{0.23\linewidth}
    \vspace{3pt}
    \centerline{\includegraphics[width=\textwidth, viewport=850 350 2500 1600,clip]{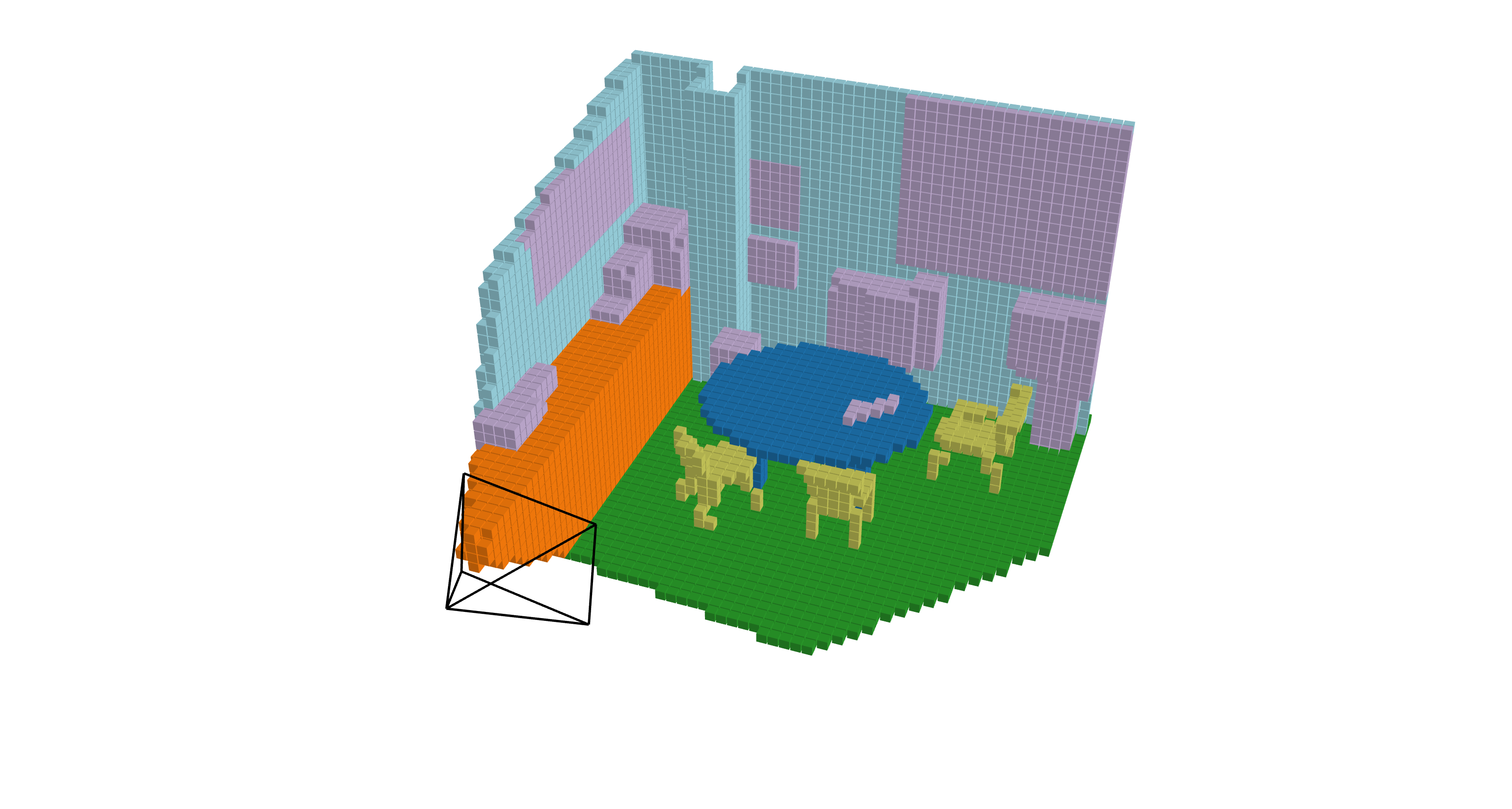}}
\end{minipage}

\begin{minipage}{0.04\linewidth}
    \vspace{3pt}
    \rotatebox{90}{Mask}
\end{minipage}
\begin{minipage}{0.23\linewidth}
    \vspace{3pt}
    \centerline{\includegraphics[width=\textwidth]{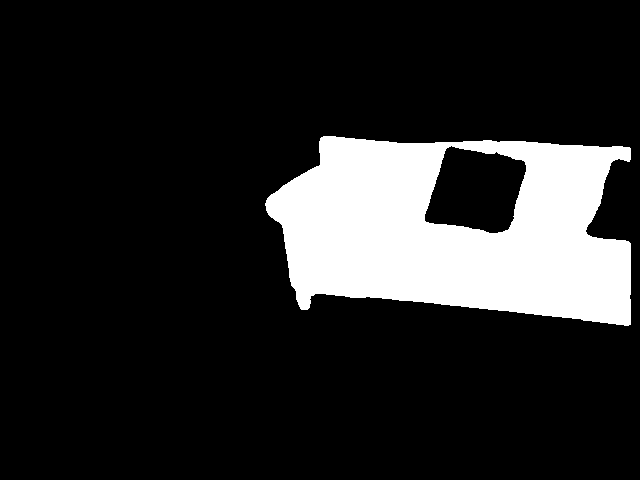}}
\end{minipage}
\begin{minipage}{0.23\linewidth}
    \vspace{3pt}
    \centerline{\includegraphics[width=\textwidth]{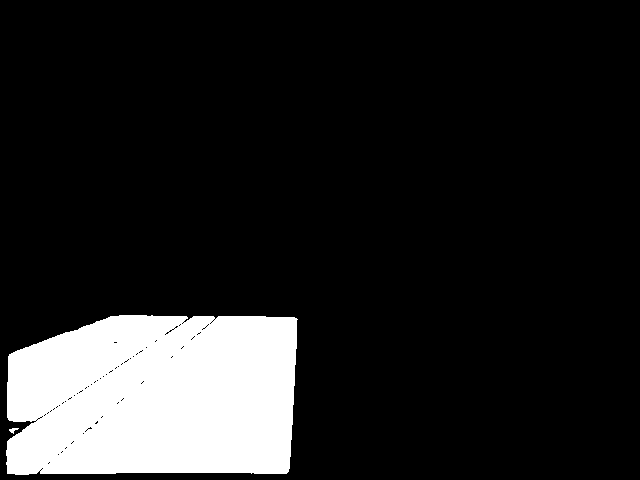}}
\end{minipage}
\begin{minipage}{0.23\linewidth}
    \vspace{3pt}
    \centerline{\includegraphics[width=\textwidth]{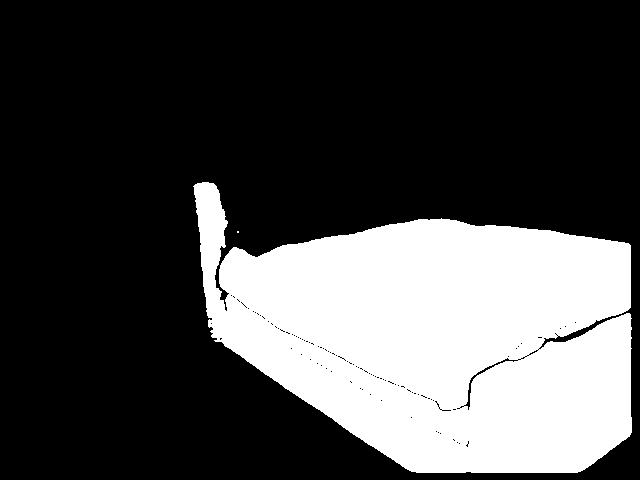}}
\end{minipage}
\begin{minipage}{0.23\linewidth}
    \vspace{3pt}
    \centerline{\includegraphics[width=\textwidth]{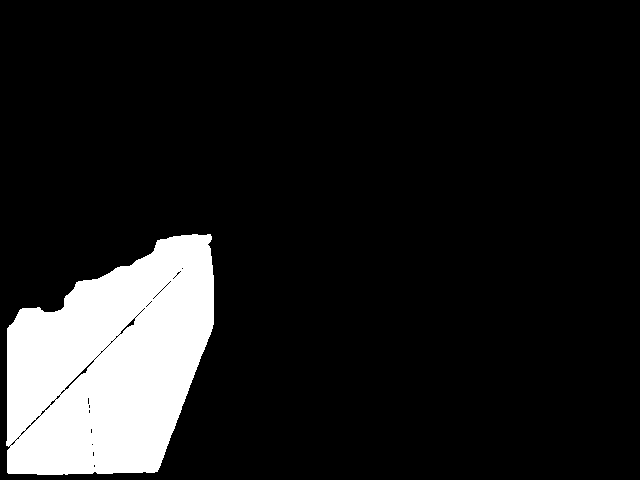}}
\end{minipage}

\begin{minipage}{0.04\linewidth}
    \vspace{3pt}
    \rotatebox{90}{Target Instance}
\end{minipage}
\begin{minipage}{0.23\linewidth}
    \vspace{3pt}
    \centerline{\includegraphics[width=\textwidth, viewport=1100 300 2300 1500,clip]{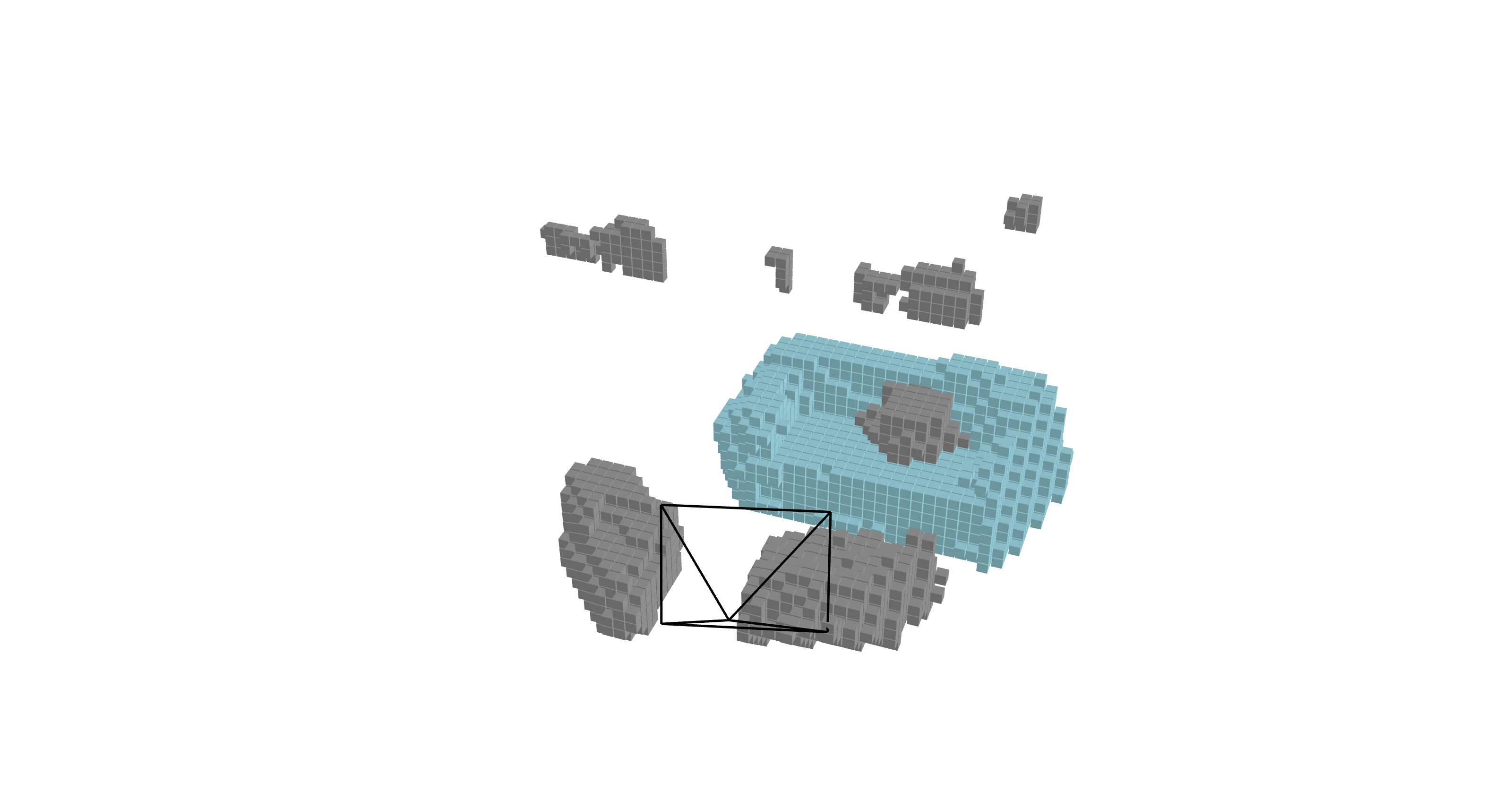}}
\end{minipage}
\begin{minipage}{0.23\linewidth}
    \vspace{3pt}
    \centerline{\includegraphics[width=\textwidth, viewport=500 350 1700 1400,clip]{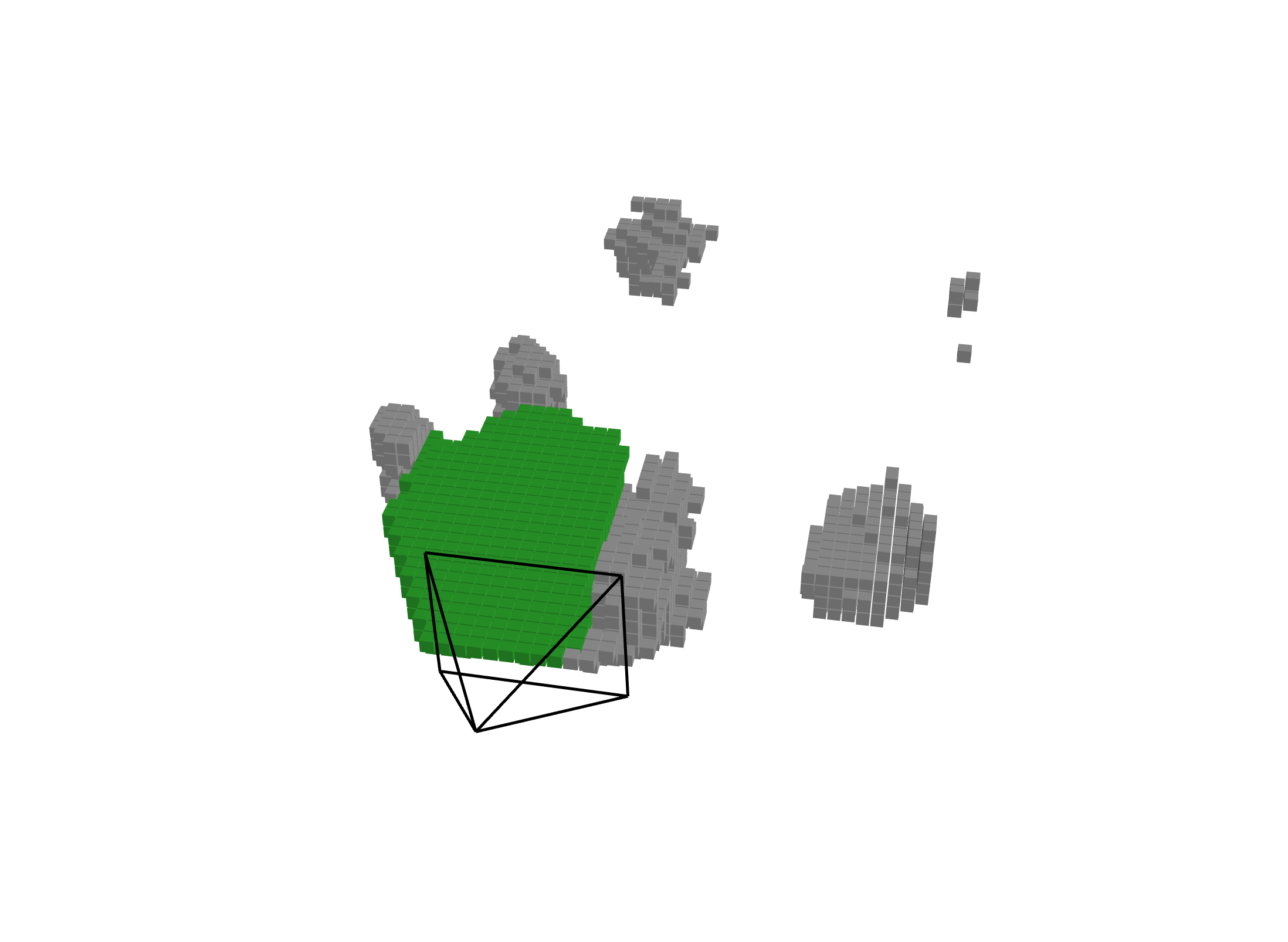}}
\end{minipage}
\begin{minipage}{0.23\linewidth}
    \vspace{3pt}
    \centerline{\includegraphics[width=\textwidth, viewport=600 150 2200 1600,clip]{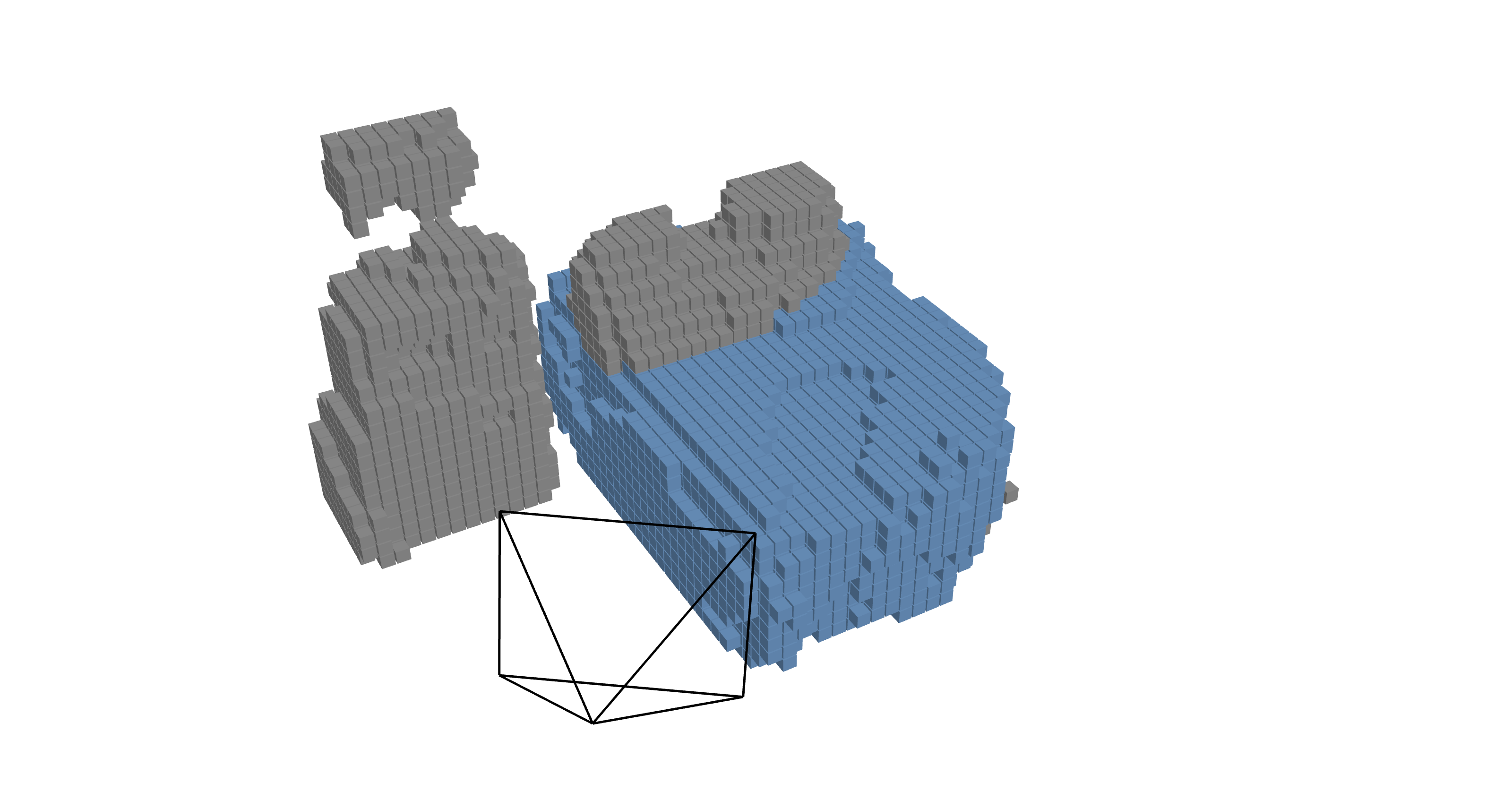}}
\end{minipage}
\begin{minipage}{0.23\linewidth}
    \vspace{3pt}
    \centerline{\includegraphics[width=\textwidth, viewport=850 300 2500 1600,clip]{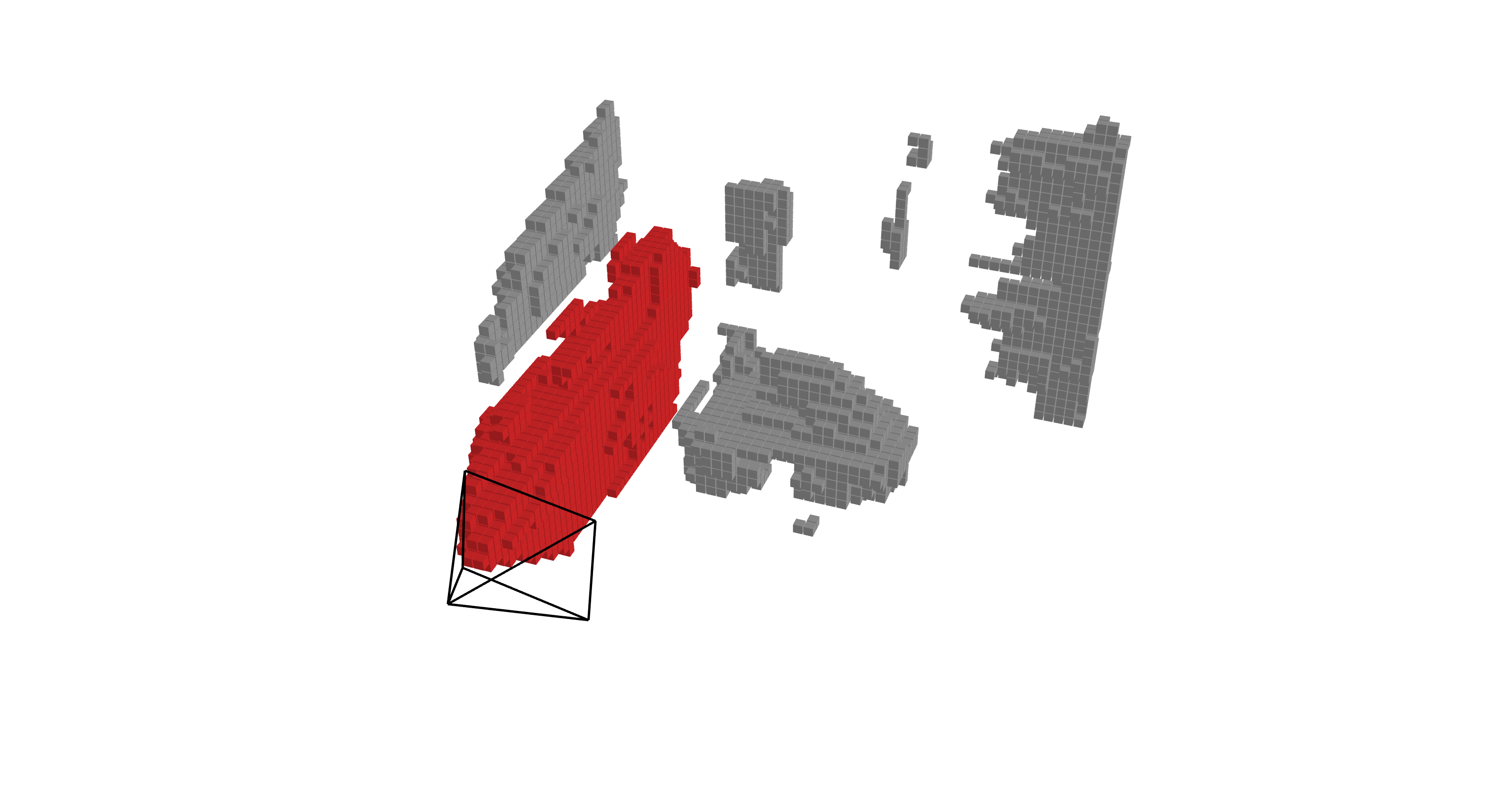}}
\end{minipage}

\caption{\small \textbf{Qualitative visualization on NYUv2 dataset}. We filter out "ceiling", "floor" and "wall" in our results. The colors of the voxels, except GT, do not indicate semantic information, but only instance segmentation results. }
\label{fig:exp}
\end{figure*}

The qualitative results of OG is shown in Fig.~\ref{fig:exp}. We can see that OG has good instance segmentation capability and can integrate Grounded-SAM to get instance segmentation results interactively.

\section{Conclusion and discussion}
Instance segmentation has been proven as an important task in 2D computer vision. Go a step further, in this paper we proposed instance segmentation in occupancy detection. Without bells and whistles, we equip naive occupancy pipeline with instance segmentation ability by simple affinity field. With the essential help above, occupancy grounding could be solved for the first time. However, our pipeline is still relatively heavy by adding extra Grounded-SAM module. Future work may investigate fusing grounding module into occupancy detection model.

\bibliography{references}
\bibliographystyle{plain}

\end{document}